\documentclass[preprint,10pt,5p]{elsarticle}
   
\usepackage{amssymb}

\usepackage{lineno}
\usepackage{amsmath}

\usepackage{amsmath}
\usepackage{algorithm}
\usepackage[noend]{algpseudocode}

\usepackage{subcaption}

\usepackage{hhline}

\usepackage{hyperref}

\usepackage{booktabs}
\usepackage{textcomp}

\usepackage[table]{xcolor}
\usepackage{colortbl}

\journal{Computers and Electronics in Agriculture}

\begin{document}

\begin{frontmatter}

\title{Pig aggression classification using CNN, Transformers and Recurrent Networks}

\address[label1]{Universidade Católica Dom Bosco, Campo Grande, Brazil}
\address[label2]{Universidade Federal de Mato Grosso do Sul, Campo Grande, Brazil}

\author[label2]{Junior Silva Souza\corref{cor1}}
\ead{junior.souza@ufms.br}
\cortext[cor1]{Corresponding author}

\author[label1]{Eduardo Bedin}
\ead{ra865659@ucdb.br}

\author[label1]{Gabriel Toshio Hirokawa Higa}
\ead{gabrieltoshio03@gmail.com}

\author[label1]{Newton Loebens}
\ead{ newtonloebens@gmail.com}

\author[label1,label2]{Hemerson Pistori}
\ead{pistori@ucdb.br}

\begin{abstract}


The development of techniques that can be used to analyze and detect  animal behavior is a crucial activity for the livestock sector, as it is possible to monitor the stress and animal welfare and contributes to decision making in the farm. Thus, the development of applications can assist breeders in making decisions to improve production performance  and reduce costs, once the animal behavior is analyzed by humans and this can lead to susceptible errors and time consumption. Aggressiveness in pigs is an example of behavior  that is studied to reduce its impact through animal classification and identification. However, this process is laborious and susceptible to errors, which can be reduced through  automation by visually classifying videos captured in controlled environment.  The captured videos can be used for training and, as a result, for classification through computer vision and artificial intelligence, employing neural network techniques. The main techniques utilized in this study are variants of transformers: STAM, TimeSformer, and ViViT, as well as techniques using convolutions, such as ResNet3D2, Resnet(2+1)D, and CnnLstm. These techniques were employed for pig video classification with the objective of identifying aggressive and non-aggressive behaviors.
In this work, various techniques were compared to analyze the contribution of using transformers, in addition to the effectiveness of the convolution technique in video classification. The performance was evaluated using accuracy, precision, and recall. The TimerSformer technique showed the best results in video classification,  with median accuracy  of 0.729.



\end{abstract}

\end{frontmatter}



\section{Introduction}\label{intro}

The increasing worldwide demand for protein in the last years has led pig farms to expand. This expansion has been financed by companies that aim at large scale production to supply the market. In the context of livestock farming, abnormal animal behavior stemming from stress factors due to the environment is one factor that can indicate issues and must be noticed in order to maintain high production levels \cite{lassaletta2019future}. Some of the behaviors that require attention are aggressiveness, feeding and mounting. Animal behavior monitoring is usually done manually, and require not only technical knowledge, but also experience. It is, therefore, a highly error-prone task~\cite{alameer2020automatic}. Given the many difficulties involved in this task, the application of technological solutions to monitor animal behavior has been increasing in recent years~\cite{shao2021pig}.

The development of deep learning-based computer vision techniques in the last decade has opened new possibilities for automating tasks related to many kinds of problems, including classification, detection and object tracking. Most relevantly, of course, it has also opened new possibilities for the development of applications aimed at, for example, tracking animals in groups, such as in the work of~\citet{xu2022automatic}. In swine farming, cameras can be used to capture images or videos, which can then be processed in order to identify abnormalities~\cite{ma2022basic}.

There are important differences between images and videos, as regards the data and information captured. By using videos, it is possible to obtain information about events that occur over a period of time (as opposed to events occurring in a given instant), which can indicate desirable or undesirable behavioral characteristics in swines that occurs in short time span. Among the undesirable behaviors, aggressiveness is one that cannot be easily identified by image analysis alone. In such cases, analyzing videos can yield better results by providing temporal information ~\cite{wang2022towards}.

The fundamental idea of this paper, therefore, is that aggressive behavior can be identified using both spacial information (extracted from individual frames) and temporal information (extracted from sequences of frames), that is, aggressive behavior can be classified through automated video analysis with neural networks. With this hypothesis in mind, this work evaluates video classification techniques applied to the task. More specifically, it evaluates models based on transformers for video processing: Visual Vision Transformer (ViViT), Space Time Attention Model (STAM) and TimeSformer, along with a Convolutional Neural Network (CNN) with Recurrent Neural Network (RNN), ResNet3D and Resnet(2+1)D.


The contribution of this work lies in the creation of a dataset containing videos depicting both aggressive and non-aggressive behaviors, collected within a local commercial breeding environment. Additionally, it proposes the automated classification of aggressive behavior, specifically those occurring in commercial breeding contexts, utilizing computer vision techniques such as convolution, recurrence and transformers for video classification. In this manner, a comparative analysis of the performance of these techniques is conducted, with objective of determined which among them yields superior results.


\section{Related Works}\label{works}

The application of deep learning techniques to solve computer vision problems has allowed the development of high performance techniques for tasks such as image and video classification, semantic segmentation and object detection. The application of these new technologies to animal monitoring has already been the subject of several studies. Some works, such as those by~\citet{hansen2021towards} and~\citet{shigang2021pig} have noticed the possibility of using images of pig faces to monitor variables such as welfare and mood, and also as an auxiliary management tool in general. Going beyond their field, computer vision techniques have also influenced researches such as that by~\citet{liao2022domestic}, that proposes the TransformerCNN model to monitor pigs by processing their vocal sounds. \citet{hu2021dual} studied the application of feature pyramid networks (FPNs) to segment individual pigs. When behavior is specifically concerned, there are also important works. \citet{jast} proposed a system to recognize feeding behavior using object detection networks, such as YoloV3. \citet{ZHANG2019104884} proposed an algorithm to detect drinking, urination and mounting behavior in real-time, by processing images generated in a pig pen. In these cases, the primary focus was on still images. Videos were used as a way of producing singular images more efficiently, which is a common practice, since the cameras can be left hanging over the pig pen, or attached to something, and configured to take videos from which individual frames can be sampled as pictures. 

On the other hand, \citet{zhang2020automated} used convolutional neural networks to process videos, in order to identify five behaviors of interest: sleeping, mounting, lying down, feeding and walking. The proposed approach used optical flow techniques to extract movement information, and convolutions were applied in parallel on RGB video frames. The detection and monitoring of animals was also analyzed by~\citet{fernandez2020computer}, who performed identification and monitoring by using CNN and Kalman filter for tracking and analyzing boars afflicted with African swine fever. In this last case, the deep learning techniques were used as part of another research aiming at proving the point that this sickness can be identified early by monitoring the movement of the animals. Another relevant case where deep learning was used within a research is the work by \citet{chen2020computer}, where the interaction of animals with enrichment objects was analyzed as a way of reducing aggressive behavior. Finally,\citet{wang2021pca} applied a CNN with Long Short-Term Memory (LSTM) to classify videos of piglets according to their postures. The authors utilized PCA to select frames, reducing their numbers and facilitating network training. As stated above, many works use videos to capture still images, and process still images for information. For many problems, single images are enough for classification using neural networks. However, these works show that behavior monitoring can benefit from the extraction of deeper spatial and temporal information from sequential image data.

There are recent developments in the field of computer vision that may prove useful for animal behavior monitoring, and have not yet been properly evaluated. Above all, these are represented by the family of transformer-based architectures, whose central component is their attention mechanism. The use of attention in image and video classification allow the implementation of techniques aimed at highlighting important features present in the data, along with their relationship~\cite{vaswani2017attention},~\cite{dai2021dynamic}. \citet{dosovitskiy2020image} proposed to apply attention concepts to (still) image processing. Then,~\citet{multViT} proposed to expand their application to video processing. Given the current state of arts, in this work, we present a video dataset with videos of aggressive behaviors of pigs. Furthermore, it includes an evaluation of transformer-based, recurrent, and convolutional neural techniques for video classification, specifically applied to identify instances of aggressive behavior."

\section{Materials and Methods}\label{methods}

\subsection{Video dataset}

For this research, a video dataset was collected in a private property in the state of Mato Grosso do Sul, Brazil. The dataset includes videos of pigs that were raised as livestock and were between 60 to 180 days old at time of capture. The images were collected in period from March until July of  2022, when the mean temperature was 28\textcelsius. The videos were captured mostly during the early  morning hours, between 08h00 to 10h00, and the late afternoon hours, between 13h00 to 16h00, when the animals were more active.

The dataset consists of 421 video clips, of which 143 were annotated as “Aggressive” and 278 were annotated as “Not Aggressive”. The videos were captured from a height of 2.6 meters and a distance of 5 meters from each enclosure, utilizing a fixed diagonal camera position. The camera device used was a Motorola e(6i) configured to capture frames with a height of 2340 pixels and a width of 4160 pixels, using the RGB color space at a rate of 30 frames per second, in recording periods of 39 minutes for each video, which were used to obtain the clips.



\begin{figure}[!htp]
    \centering
    \includegraphics[scale=0.35]{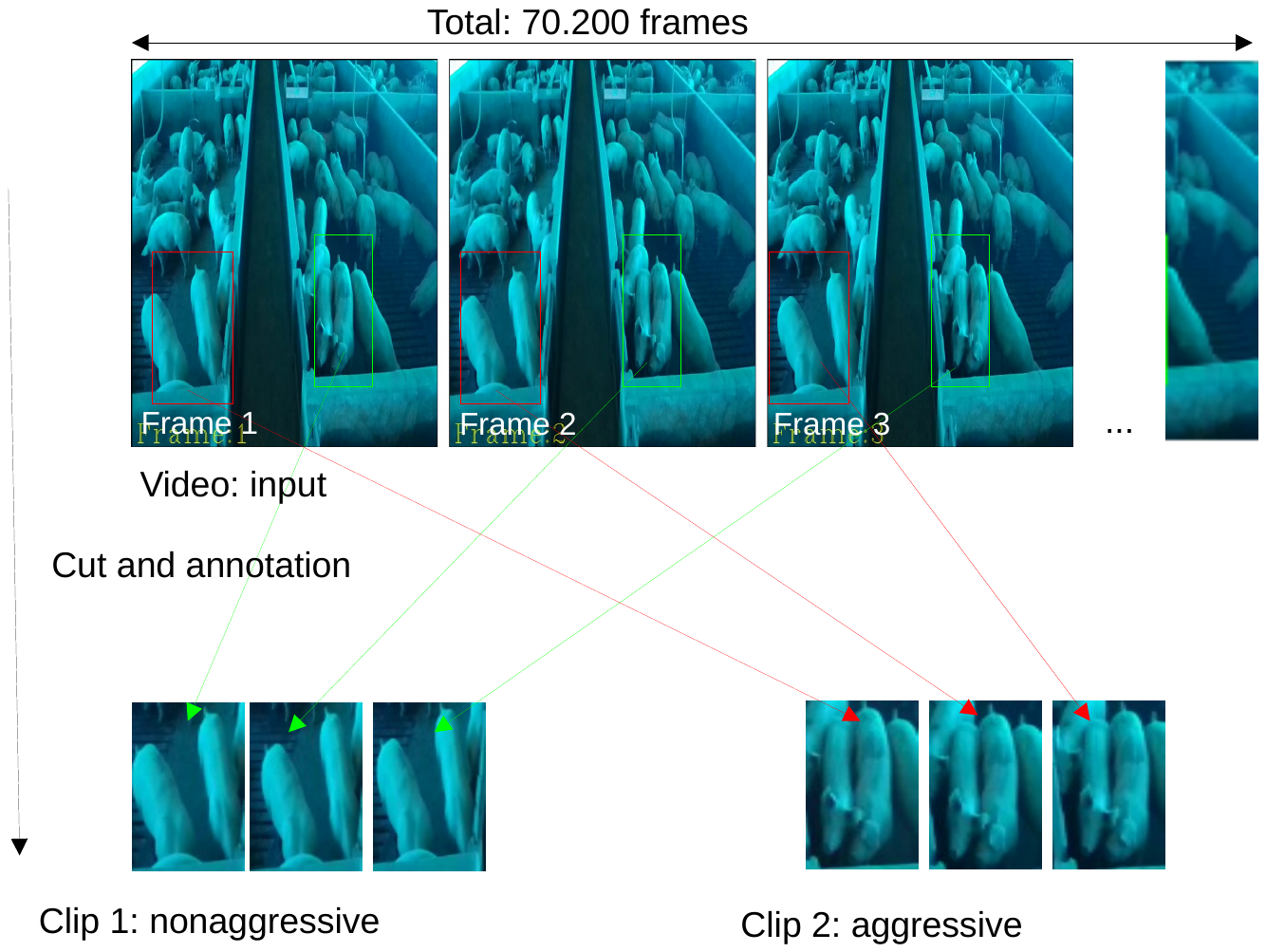}
    \caption{In this image, we can visualize the process of classification from an input video, where clips are obtained.}
    \label{fig:videoGeral}
\end{figure}

To obtain the dataset, it was necessary to perform a  processing of edition \footnote{The editing process  was performed using the python programming language  and the OpenCV package.} on each video recording. The videos were cut into clips at the moments when the animals exhibited  behavior considered aggressive, as well as moments when they did not display aggression  but were in close proximity.  In Figure ~\ref{fig:videoGeral} we can observe an example of an input video recording composed of 70,200 frames (39 minutes at 30 frames per second), where the manual classification (aggression or nonaggression) was performed. Furthermore, it is possible to observe that the image presents a blue tone due to the  barn fence on the side of the enclosure where the animals are housed. This is done to provide both light and ventilation during the day, a practice adopted by many local shelters.

\begin{figure}[!ht]
    \centering
    \includegraphics[scale=0.25]{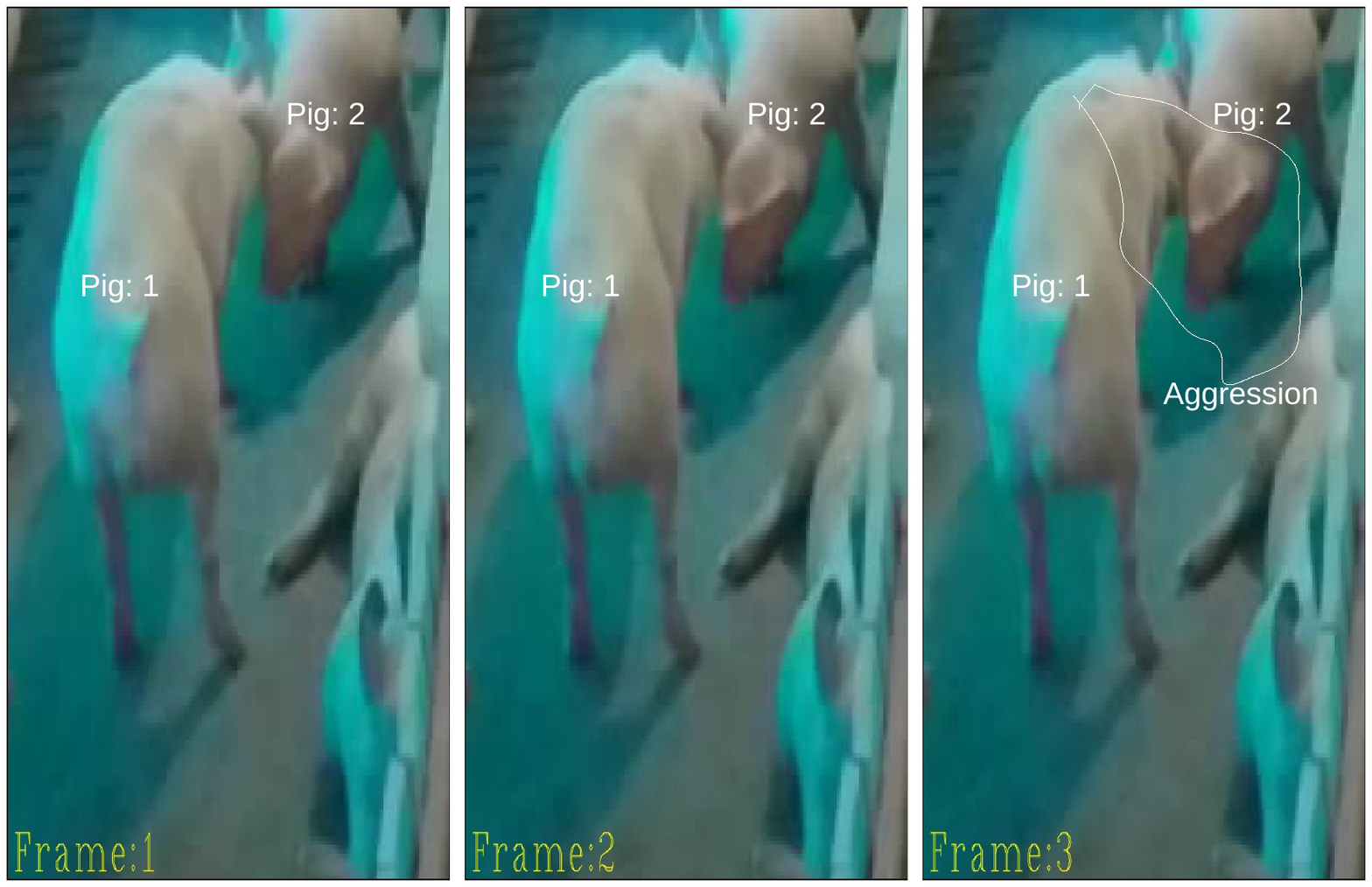}
    \caption{The image shows three sequential frames depicting  two instances of aggression. We can observe more highlighted details in frame three, located on the right, where the animals are attempting to bite.}
    \label{fig:aggressivefig}
\end{figure}

In Figure~\ref{fig:aggressivefig}, We can see an example of three frames that depict two pigs in an aggressive moment from a clip. The animals involved, pig 1 and pig 2, display an initial aggression, attempting to bite each other. This is a brief moment that occurs within a short time interval.

\begin{figure}[!ht]
    \centering
    \includegraphics[scale=0.25]{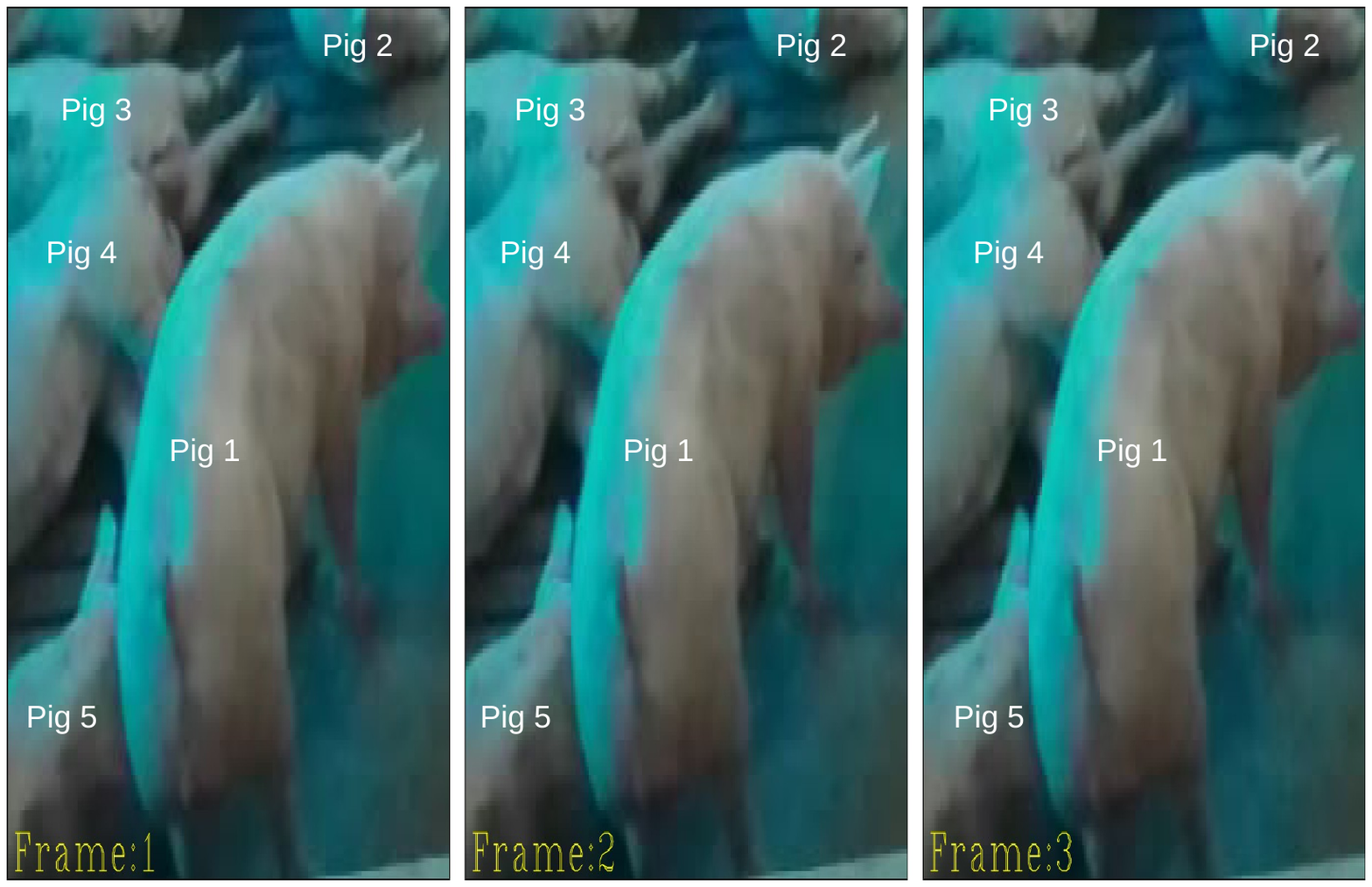}
    \caption{Image depicting three frames illustrating non-aggressive animal behavior.}
    \label{fig:nonaggressivefig}
\end{figure}

The animal behavior, considered non-aggressive, can be observed in Figure~\ref{fig:nonaggressivefig}, where we can see animals in standing up and animals lying down in normal conditions. 

\subsection{Video preprocessing}

Video compression was employed to reduce the number of frames to be processed by eliminating redundant information from the clips. In the work of \citet{liu2020computer}, the reduction of frames was analyzed to determine the  minimum and sufficient amount of frames needed for training and classification. This processing allowed for a reduction of processing during the training.


The reduction of clip frames was accomplished at a 4:1 ratio. As a result, each sequence of 40 frames was condensed to 10 frames.
Besides the compression, each frame of video was also resized to 400 pixels of height and 400 pixels in width.





\subsection{Deep Learning}


This paper compares the performance of deep neural networks applied to video classification, including  convolutional neural networks (CNNs), recurrent neural networks (RNNs), and transformers-based neural networks. Convolutional neural networks, either combined with or without recurrent neural networks, are considered as alternative approaches for video classification, as demonstrated in previous works such as ~\citet{patil2021video} and ~\citet{cnnlstmOnly}.

To perform the classification of videos using convolutional neural networks, architectures have been proposed in \citet{tran2018closer} that utilize the mechanism of  \textquotedblleft skip connection \textquotedblright to handle the problem of vanishing gradients which can reduce the magnitude of the updates  of neurons in training. In order to work with spacial and temporal information, two architectures have been developed:  ResNet3D and Resnet(2+1)D also referring to as ResNet21D in this work. ResNet3D performs convolutions in three dimensions to extract both temporal and spacial information from video frames. On the other hand, the ResNet(2+1)D architecture performs  2D convolutions followed by 1D convolutions, extracting spacial and temporal information, respectively.

The use of recurrent neural networks (RNNs) in combination with convolutional neural networks (CNNs) was also proposed in this work. The network used was ResNet. The CNN was utilized for extracting features from videos, which represent spatial information. Meanwhile, the RNN was employed for classification and extraction of information from frames, representing temporal information \cite{harris2023deepaction}.


Transformers are implementations of attention mechanisms developed initially in the field of natural language processing. They have since been adapted for use in computer vision in tasks such as image and video classification, detection and segmentation\cite{9574462Bi}.


In natural language processing, each word is represented in the form of tokens. However, in computer vision, an image is divided in small parts called patches, which are used as tokens, and are them used for image classification using ViT (Vision Transformers).


The ViT introduces the construction of a neural network that uses transformers, as an adaptation that does not use convolutions \cite{dosovitskiy2020image, vitOnly}. The tokens are extracted from an image and passed to a transformer encoder which corresponds to one block of the transformer. One of the differentials of the transformer is its attention mechanism, specifically the dot product attention presented in the encoder. The encoders are responsible for extracting information, which is used by a multilayer perceptron that is responsible for classification.


The ViViT architecture is an application of ViT in video and image classification problems. In this architecture, each video is treated as a sequence of images from which important spatial and temporal information are extracted for classification purposes. As videos are  composed of sequential images (frames), relevant information can be extracted at each frame change. Furthermore, each frame in the video can be analyzed by ViT, which extracts spatial information that is used to extract temporal information. The architecture ViViT has two stages of application using ViT: The first stage is applied to each frame from the video and the second stage is applied to the results from the first stage  \cite{Arnab_2021_ICCV}.


The STAM architecture is similar to ViViT. This architecture applies  ViT to extract spatial and temporal information, which gives more efficient processing while  prioritizing the extraction of spatial information before capturing temporal information. In STAM, 16 frames are selected per video for processing, as described in \cite{stamOnly}.


The TimeSformer (Time-Space Transformer) architecture divides each frame of the video into patches with a size of 16x16, which can be changed to other sizes such as 32x32, 64x64. This architecture uses ViT  to extract spatial and temporal information from the obtained patches. During processing, only the patches with more information are selected. The quantity and location of patches can vary during the processing, which involves the execution of multiple  ViT models. This processing allows the model to focus on the most  important parts of each frame to achieve  video classification \cite{timeSformerOnly}.

\subsection{Experimental Setup}


To conduct the experiments, the dataset was used with a 5-fold cross-validation ~\cite{liu2019margin}. In each run, 20\% of the training images were used for validation. The frames of the videos were resized to $(64,64)$ pixels due to the limitations of the runtime environment, which used the NVIDIA TITAN Xp with 12 GB of memory, supporting execution with sizes no larger than that. For the \textit{transformer} modules, 32 patches were extracted from each frame. This initial value is used in works such as \cite{Strudel2021ICCV}. Each architecture was tested with the following learning rates: 0.0001 and 0.0003, values utilized in different works, such as that of \citet{islam2020human}, as well as that of \citet{s19173738}. In this work, the neural networks were optimized with Adam, which yielded good results in works such as that of \citet{desai2020comparative} and \citet{almadani}.

The training was performed for 20 epochs, with batches of size 8, following the approach by \citet{10042233Zhou}. The weights were randomly initialized, and the transfer learning technique was not utilized due to the specific characteristics of the application domain. At the end of the training phase, four classification metrics were calculated for the test folds, and the mean values were obtained for the following metrics: accuracy, precision, recall, and F-score. Then, these results were evaluated through an ANOVA hypothesis test, with a significance threshold of 5\%. The Scott-Knott clustering test was used post-hoc, when the ANOVA results were significant.





\section{Results}\label{results}


The Tables~\ref{tab:accuracy},~\ref{tab:precision},~\ref{tab:recall}, and ~\ref{tab:fmeasure} show the values of accuracy, precision, recall, and F-score, respectively, using techniques for video classification, analyzed with one-way ANOVA.  The Table columns show the name of the techniques, as well as the mean, median, and standard deviation (SD) values. 

\begin{table}[ht]
\centering
\begin{tabular}{|c|c|c|}
  \hline
  \multicolumn{3}{|c|}{Accuracy metric} \\
  \hline
    Technique & Median & Mean (SD) \\ 
  \hline
   ViViT       & 0.694         & 0.698          (0.056)          \\ 
   STAM        & 0.659         & 0.665          (0.016)          \\ 
   TimeSformer & \textbf{0.729}& \textbf{0.764} (\textbf{0.083}) \\ 
   CnnLstm     & 0.699         & 0.724          (0.092)          \\ 
   Resnet3D    & 0.724         & 0.750          (0.107)          \\ 
   Resnet21D   & 0.718         & 0.757          (0.111)          \\ 
  \hline
\end{tabular}
 \caption{The results are presented for all techniques using the accuracy metric.}
 \label{tab:accuracy}
\end{table}

\begin{table}[!ht]
\centering
\begin{tabular}{|c|c|c|}
  \hline
    \multicolumn{3}{|c|}{Precision metric} \\
  \hline
  Technique (group) & Median & Mean (SD) \\ 
  \hline
    ViViT       (a)  & 0.594          & 0.580         (0.111)          \\ 
    STAM        (b) & 0.167          & 0.300          (0.358)          \\ 
    TimeSformer (a) & \textbf{0.681} & \textbf{0.716} (\textbf{0.125}) \\ 
    CnnLstm     (a) & 0.576          & 0.614          (0.142)          \\ 
    Resnet3D    (a) & 0.613          & 0.670          (0.197)          \\ 
    Resnet21D   (a) & 0.600          & 0.653          (0.183)          \\ 
  \hline
\end{tabular}
  \caption{The results are presented for all techniques using the precision metric.}
\label{tab:precision}
\end{table}

\begin{table}[!ht]
\centering
\begin{tabular}{|c|c|c|}
  \hline
    \multicolumn{3}{|c|}{Recall metric} \\
  \hline
  Technique (group) & Median & Mean (SD) \\ 
  \hline
    ViViT       (a) & 0.358          & 0.381          (0.129)          \\ 
    STAM        (b) & 0.017          & 0.035          (0.046)          \\ 
    TimeSformer (a) & 0.483          & 0.499          (0.207)          \\ 
    CnnLstm     (a) & 0.397          & 0.480          (0.189)          \\ 
    Resnet3D    (a) & 0.534          & 0.531          (0.187)          \\ 
    Resnet21D   (a) & \textbf{0.677} & \textbf{0.613} (\textbf{0.225}) \\ 
  \hline
\end{tabular}
     \caption{The results are presented for all techniques using the recall metric.}
\label{tab:recall}
\end{table}

\begin{table}[!ht]
\centering
\begin{tabular}{|c|c|c|}
  \hline
    \multicolumn{3}{|c|}{F-score metric} \\
  \hline
   Technique (group) & Median & Mean (SD) \\ 
  \hline
    ViViT       (a) & 0.426          & 0.454          (0.125)          \\ 
    STAM        (b) & 0.031          & 0.062          (0.082)          \\ 
    TimeSformer (a) & 0.544          & 0.573          (0.178)          \\ 
    CnnLstm     (a) & 0.465          & 0.532          (0.168)          \\ 
    Resnet3D    (a) & 0.555          & 0.585          (0.185)          \\ 
    Resnet21D   (a) & \textbf{0.602} & \textbf{0.621} (\textbf{0.193}) \\ 
   \hline
\end{tabular}
     \caption{The results are presented for all techniques using the F-score metric.}
\label{tab:fmeasure}
\end{table}

The TimeSformer technique presented the best performance in the accuracy and precision metrics, with values of 0.729 and 0.764 for their respective medians and means. However, the Resnet21D technique outperformed TimeSformer in the recall metric.

The ANOVA results were marginally significant for the techniques in all four metrics, with p-values of $0.0771$, $0.000449$, $1.2 \times 10^{-8}$, and $4.9 \times 10^{-10}$ for accuracy, precision, recall, and F-score, respectively. These values were analyzed and indicate statistically significant differences between almost all analyzed techniques, except for accuracy when considering a significance level of 5\%.
Subsequent post-hoc analysis using the Scott-Knott clustering test revealed that TimesFormer outperformed the other techniques in precision metrics, despite being clustered together with ViViT, Resnet3D, Resnet21D and CnnLstm. STAM achieved the worst performance in all metrics.

The significant differences between the techniques can be analyzed using the precision metric, which yielded a p-value  of $0.000009$. These values indicate strong statistical differences  among the adopted techniques. Furthermore, when comparing the learning rate, the precision and recall metrics also showed significant differences, with p-values of  $0.049393$ and $0.01480$, respectively.

Considering the metrics of precision and a learning rate of $0.0001$, the technique that showed the best result is TimeSformer, which is a variant of  transformer model. 

We can visualize in Table~\ref{tab:classification1} six video clips were used, with each clips displaying  three frames as an illustrative example. The GT  (Ground Truth) column represents the actual outcomes, while the DL (Deep Learning) column reflects the results from TimeSformer classification technique. The clips identified as 'a' and 'b' were correctly classified as 'NA' (Non-Aggressive), whereas the clips 'c' and 'd' were accurately classified as 'A' (Aggressive). Additionally, in Table~\ref{tab:classification1}, we can observe the clips 'e' and 'f', which, although  categorized as 'NA' and 'A', respectively, in the GT column,  were not classified correctly.

\begin{table}[!ht]
  \centering
  \begin{tabular}{|c|c|c|}
    \hline
    \textbf{Frames} & \multicolumn{2}{c}{\textbf{Results}}  \\ \hline
    \hline
                    & \textbf{GT} & \textbf{DL} \\ \hline
    \hline
    (a) \includegraphics[scale=0.06]{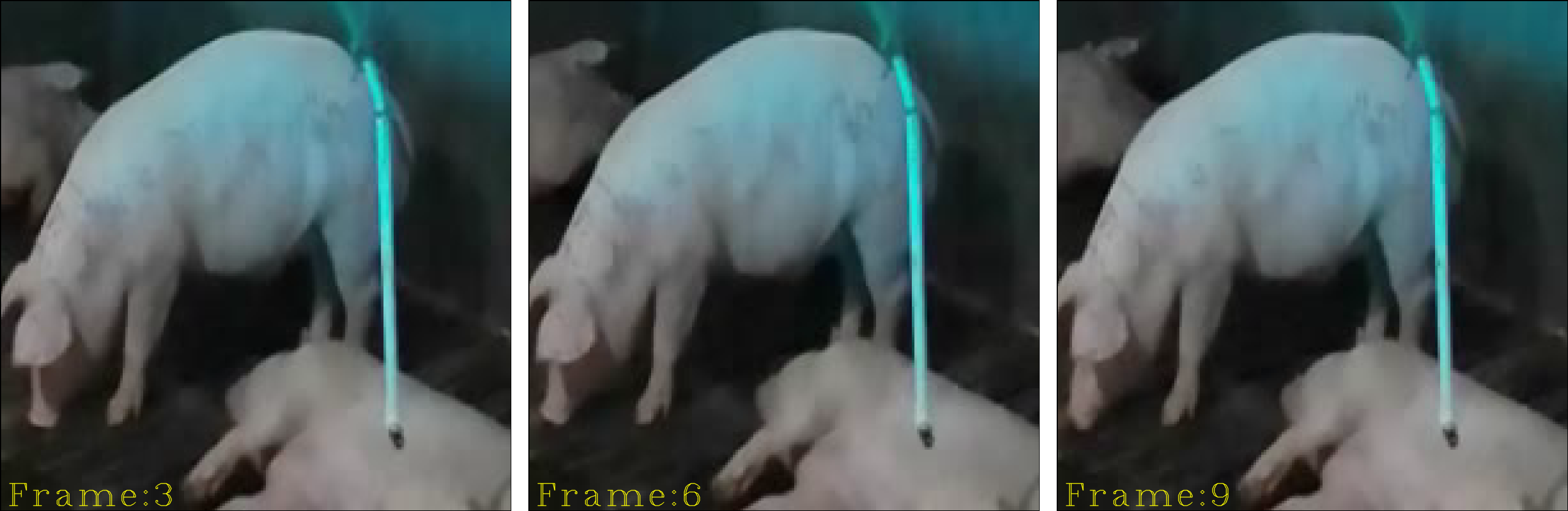} & NA & NA  \\ \hline
    (b) \includegraphics[scale=0.06]{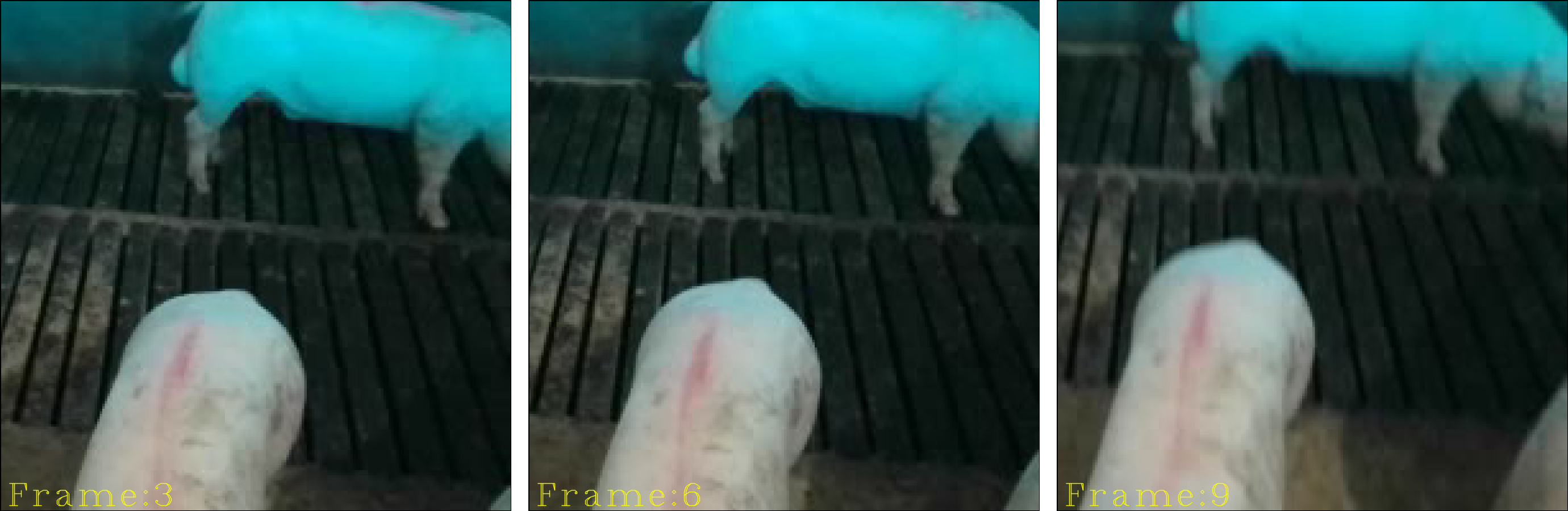} & NA & NA \\ \hline
    (c) \includegraphics[scale=0.06]{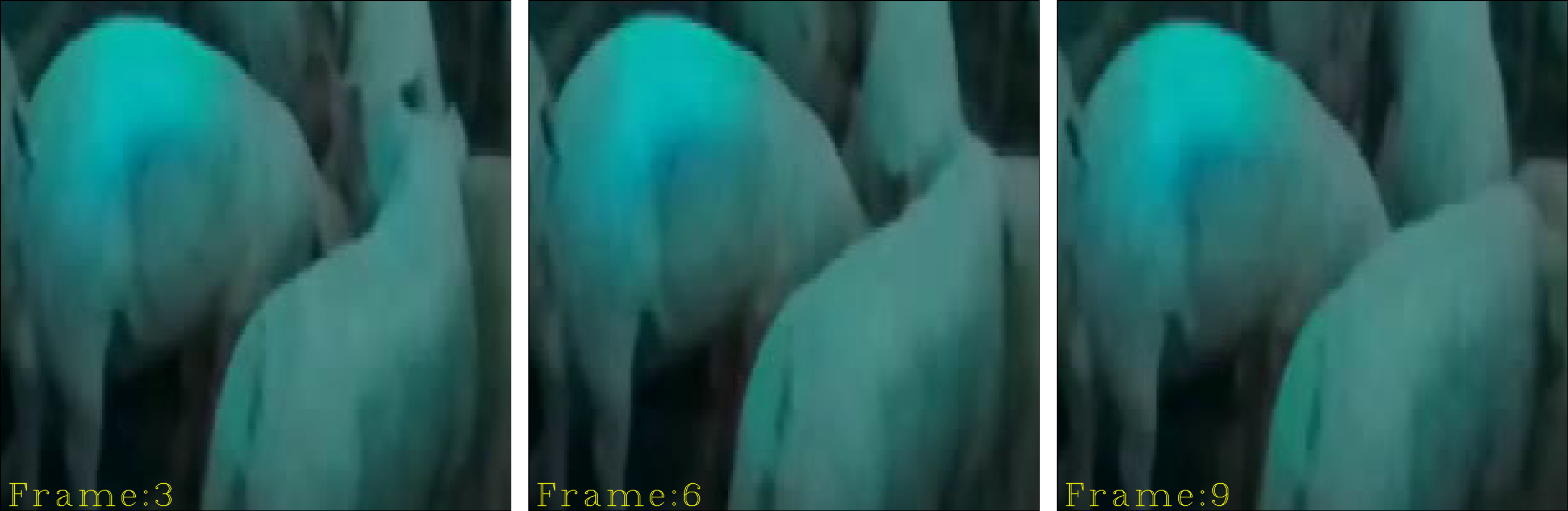} & A & A  \\ \hline
    (d) \includegraphics[scale=0.06]{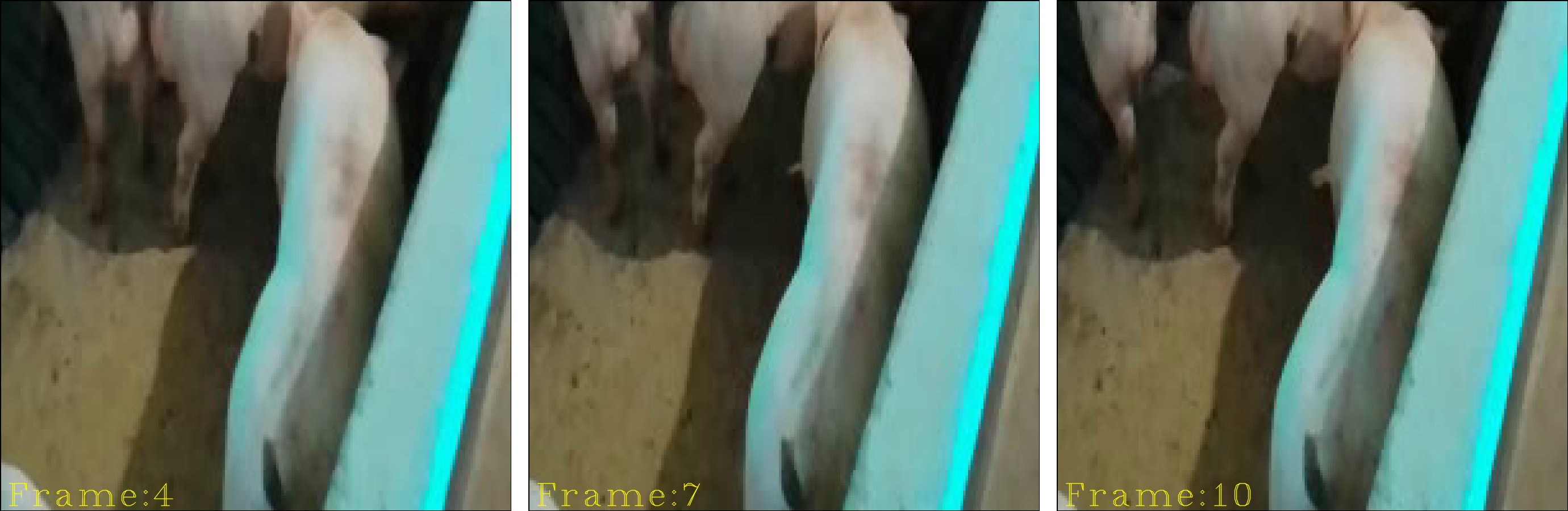} & A & A  \\ \hline
    (e) \includegraphics[scale=0.06]{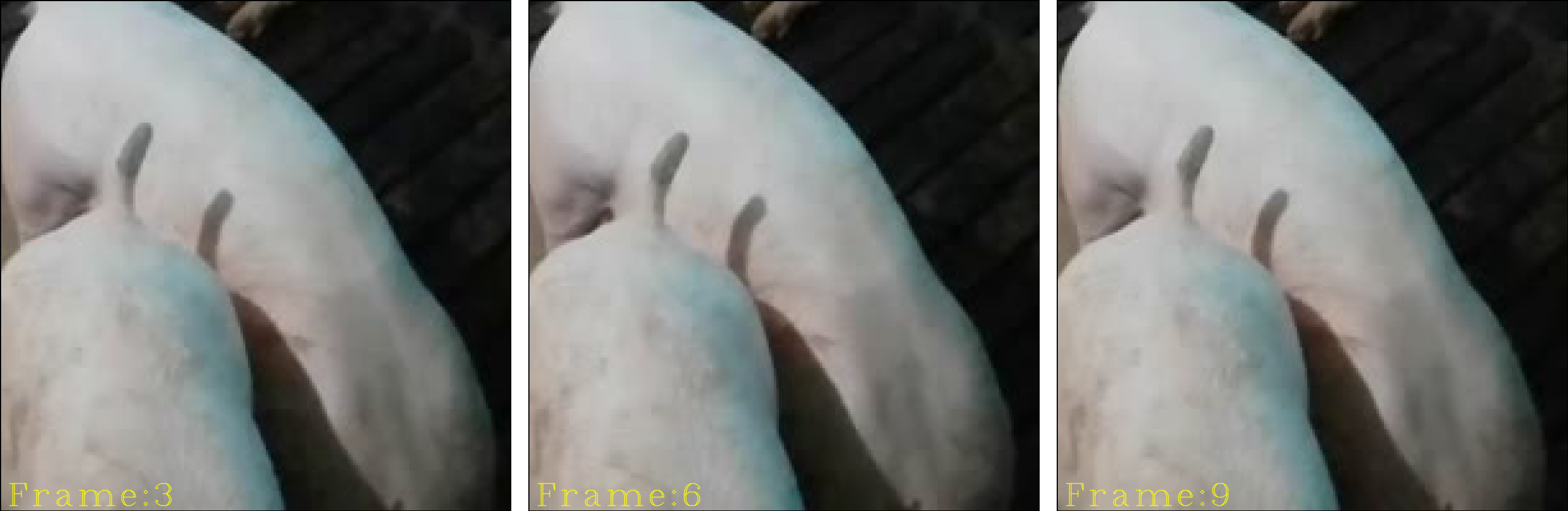} & NA & A  \\ \hline
    (f) \includegraphics[scale=0.06]{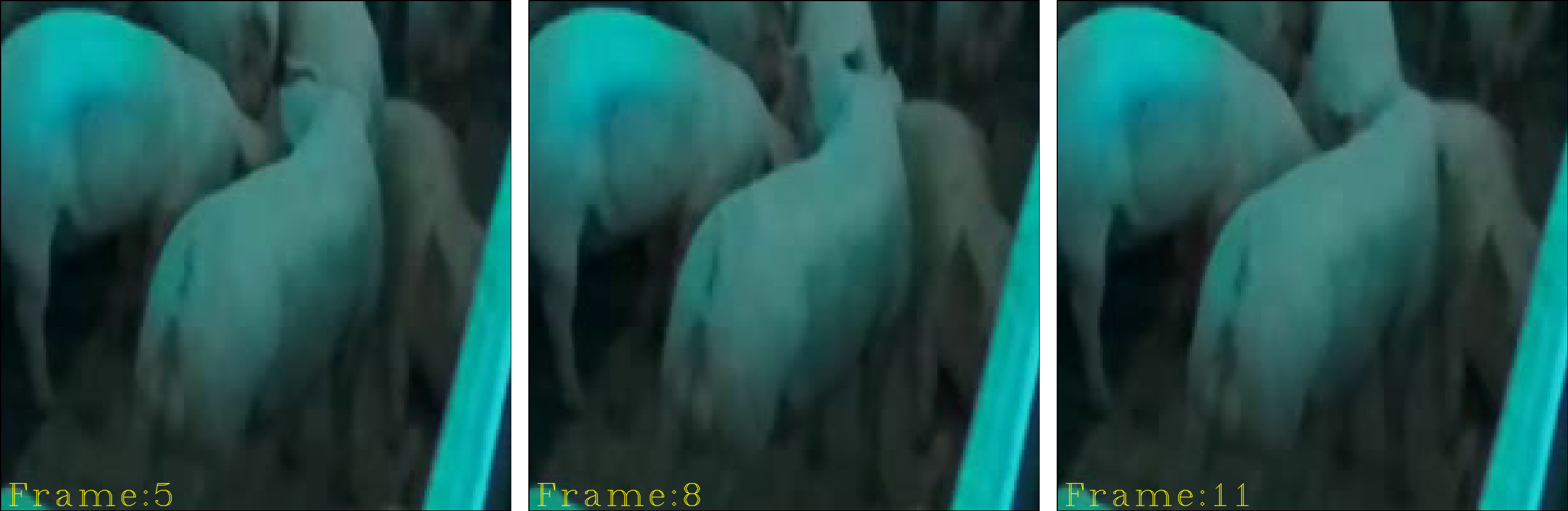} & A & NA  \\ \hline
  \end{tabular}
  \caption{Accurate and inaccurate classification of videos along with their corresponding labels. Taking into account NA (Non-aggressive), A (Aggressive), GT (Ground Truth), and DL (Deep Learning techniques).}
  \label{tab:classification1}
\end{table}

\begin{figure}[!ht]
    \centering
    \includegraphics[scale=0.65]{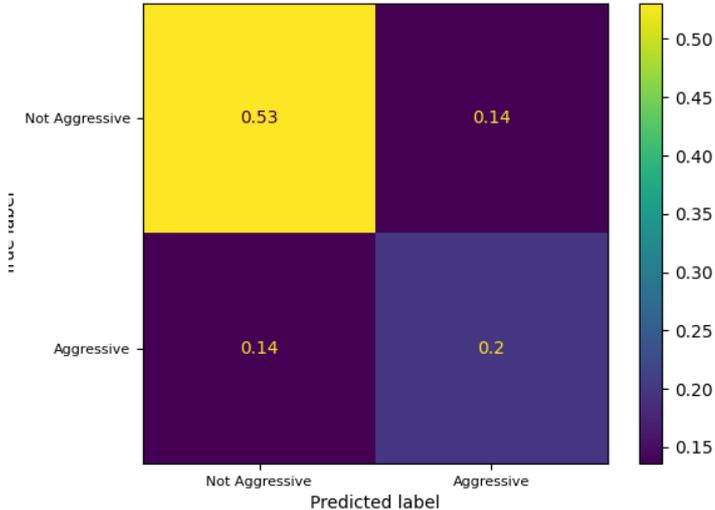}
    \caption{The confusion matrix was obtained from the testing set using the TimeSformer technique.}
    \label{fig:confusionMatrizTimeSformer}
\end{figure}

The confusion matrix was obtained for the TimeSformer technique in the testing dataset, as shown in Figure~\ref{fig:confusionMatrizTimeSformer}.
We can observe an average accuracy rate of 73\% in the classification, where there are more videos classified as non-aggressive.

\section{Discussion}

Video classification requires the extraction of temporal and spatial information that can be explored by techniques using sequences of frames (videos). In this research, we propose the use of variants of transformers for video classification. Although the  transformer technique has been applied in works involving videos, such as \cite{ZHANG2023100394}, and has shown better results when using datasets with videos of gastrointestinal diseases. The use of transformers in computer vision poses a challenge, transformers rely on extensive datasets to yield improved results, necessitating high-quality images and substantial processing power \cite{KhanSalman2022TiVA}.

The reduction of the number of processed frames is an interesting strategy  adopted  to improve processing efficiency and reduce redundant information. However, it can affect the model's generalization capacity. In the case  of STAM, which was configured to process 16 frames per video, it did not result in better precision, recall and F-score. This may be due to the fact that the behavior of interest is not always clearly presented in the initial frames of the analyzed videos.

In the case of ViViT, which is another variant of transformer-based models, it did not show better results compared by the analyzed metrics. This could  be attributed to differences in architecture and the way ViViT processes and extracts information from  frames. On the other hand, TimeSformer emerged as a better option, particularly in terms of precision. Its approach of selecting relevant patches  from each frame allows for reducing redundant information and focusing on the most important characteristics for classification. This strategy of patch selection proves advantageous for the classification task and leads to superior results, especially in terms of precision.

Although the use of CnnLstm with recurrent neural networks has shown better results in previous works, such as \cite{han2023evaluation}, this  technique did not demonstrate the best performance in this project. This can be attributed to the fact  that the analyzed videos consist of series of frames that require the processing of long sequences, which becomes a bottleneck for an LSTM-based to approach.

However, it is important emphasize that the other techniques that exclusively used  convolution, such as Resnet3D and Resnet21D, achieved better results compared to the CNN-LSTM approach. This indicates that combining LSTM with convolution does not  yield significant improvements in this specific context.


\section{Conclusion}\label{conclusion}


This work proposed the use  of deep learning techniques, including transformers, convolutional neural networks, and recurrent layers to classify aggressive behavior in pigs. To accomplish this, a video dataset was created for training and validating the employed techniques.


The techniques were evaluated using accuracy, precision, recall, and F-measure metrics, which demonstrated that the TimeSformer technique achieved the best results in video classification across all metrics. While variants of transformers have been utilized in numerous works involving video classification, the TimeSformer technique outperformed them with the dataset created and used in this study, particularly in terms of precision.








\end{document}